\documentclass[10pt,twocolumn,letterpaper]{article}

\usepackage{cvpr}
\usepackage{times}
\usepackage{epsfig}
\usepackage{graphicx}
\usepackage{amsmath}
\usepackage{amssymb}
\usepackage{algorithm, algorithmicx}
\usepackage[noend]{algpseudocode}

\usepackage{algpseudocode}
\usepackage{textcomp}
\usepackage{xcolor}
\usepackage{longtable}
\usepackage{booktabs}
\usepackage{multirow}
\usepackage{tabu}
\usepackage{authblk}

\usepackage[breaklinks=true,bookmarks=false]{hyperref}

\cvprfinalcopy 

\def\cvprPaperID{****} 

\begin{document}

\title{Learning Scene Structure Guidance via Cross-Task Knowledge Transfer for Single Depth Super-Resolution}

\author{Baoli~Sun$^{1}$, Xinchen~Ye$^{1, 2}$\thanks{Corresponding author: yexch@dlut.edu.cn.  This work was supported by National Natural Science Foundation of China (NSFC) under Grant 61702078, 61772108, 61976038, 61772106.}, Baopu~Li$^{3}$, Haojie~Li$^{1, 2}$, Zhihui~Wang$^{1, 2}$,  Rui~Xu$^{1, 2}$}
\affil{$^{1}$International School of Information Science $\&$ Engineering, Dalian University of Technology, China\\ $^{2}$Key Laboratory for Ubiquitous Network and Service Software of Liaoning Province, China\\    $^3$ Baidu Research, USA \\}

\maketitle
\thispagestyle{empty}
\pagestyle{empty}

\begin{abstract}
Existing color-guided depth super-resolution (DSR) approaches require paired RGB-D data as training samples where the RGB image is used as structural guidance to recover the degraded depth map due to their geometrical similarity. However, the paired data may be limited or expensive to be collected in actual testing environment.
Therefore, we explore for the first time to learn the cross-modality knowledge at training stage, where both RGB and depth modalities are available, but test on the target dataset, where only single depth modality exists. Our key idea is to distill the knowledge of scene structural guidance from RGB modality to the single DSR task without changing its network architecture. Specifically, we construct an auxiliary depth estimation (DE) task that takes an RGB image as input to estimate a depth map, and train both DSR task and DE task collaboratively to boost the performance of DSR. Upon this, a cross-task interaction module is proposed to realize bilateral cross-task knowledge transfer. First, we design a cross-task distillation scheme that encourages DSR and DE
networks to learn from each other in a teacher-student role-exchanging fashion. Then, we advance a structure prediction (SP) task that provides extra structure regularization to help both DSR and DE networks learn more informative structure representations for depth recovery.  Extensive experiments demonstrate that our scheme achieves superior performance in comparison with other DSR methods. 
\end{abstract}

\section{Introduction}
To better understand a scene image, depth information is supplemented to the RGB images, providing the key clue about the scene and enabling wide applications in 3D reconstruction \cite{DBLP:conf/ivs/GeigerZS11}, autonomous navigation \cite{DBLP:conf/iros/KerlSC13}, monitoring \cite{DBLP:journals/corr/abs-1812-05831}, and so on. However, acquiring depth information for indoor and outdoor scenes needs expensive cost and great efforts, especially for high-quality and high-resolution (HR) depth maps. As such, one of the effective post processing techniques, Depth Super-Resolution (DSR), is greatly desired to yield HR depth maps to alleviate this problem. Many efforts have been taken along the direction of DSR. Usually, fine scene structures are easily lost or severely destroyed in low-resolution (LR) depth map because of the limited spatial resolution. An RGB image and its associated depth map are the photometric and geometrical representations of the same scene, and have a strong structural similarity. Most existing DSR methods learn structural complementarity from RGB images to recover the degraded depth maps.

\begin{figure*}[!]
	\centering
	\centerline{\includegraphics[width=0.97\linewidth]{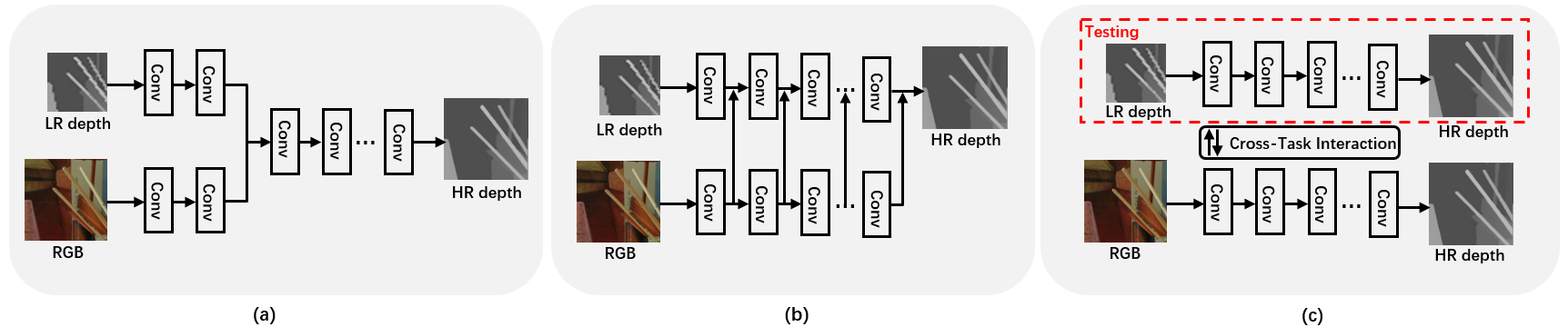}}
	\caption{Color-guided DSR paradigms. (a) Joint filtering, (b) Multi-scale feature aggregation, (c) Our cross-task interaction mechanism to distill knowledge from RGB image to DSR task without changing its network architecture.}
	\label{fig:intro2}
\vspace{-4mm}
\end{figure*}

Previous color-guided DSR methods take advantage of RGB-D image pairs via a two-way fusion architecture, in which an extra branch is required to extract structural guidance from RGB image. As illustrated in Figure \ref{fig:intro2} (a), RGB image and LR depth map are often processed by separate branches and filtered together through a joint branch to output the HR result \cite{DBLP:conf/eccv/LiHA016,DBLP:conf/cvpr/PanDRLT019,DBLP:conf/iccv/LutioDWS19,DBLP:journals/corr/abs-1903-11286}.  
However, due to the simple feature aggregation at a specific layer in the middle of the network, high-frequency structure information from RGB image is more likely to be lost in the process of feature extraction. Therefore, as shown in Figure \ref{fig:intro2} (b), some novel methods
\cite{DBLP:conf/eccv/HuiLT16, DBLP:journals/tip/GuoLGCFH19,DBLP:journals/tip/YeSWYXLL20} incorporate a new paradigm of feature aggregation, i.e, multi-scale fusion, to allow the network to learn rich hierarchical features at different levels. This in turn makes the network to retain more spatial details for recovering both fine-scale and large-scale structures.

Although existing color-guided DSR methods have demonstrated remarkable progress, several limitations still remain. First, these methods require paired RGB-D data as training examples to jointly recover the degraded depth map. However, the paired data may be limited or expensive to be collected in actual testing environment. For example, RGB image and depth map are captured by separate depth and RGB sensors with different resolutions and views, thus needing accurate calibration and rectification between them to obtain the registered pairs. Actually, most of real-world applications still come with only a single LR depth map, which raises the above question.
Second, considering the memory consumption and computing burden, the processing on the HR RGB data also hinders the practical application. 
Moreover, although RGB features can be used as structural guidance to resolve the degradation in DSR, RGB discontinuities do not always coincide with those of depth map (structure inconsistency), which results in noticeable artifacts such as texture copying and depth bleeding. Therefore, how to leverage RGB information to help recover the depth map and simultaneously satisfy the actual testing environment, still needs to be studied.

Motivated by  the above analysis, this paper breaks away from the shackles of general paradigms and introduces a novel scene structure guidance learning method for the task of DSR, as shown in Figure \ref{fig:intro2} (c).
We explore for the first time to learn the cross-modality knowledge at training stage, where both RGB and depth modalities are available, but test on the target dataset, where only single depth modality exists. Our key idea is to distill the knowledge of scene structural guidance from RGB modality to the single DSR task without changing its network architecture.

\begin{figure*}[ht]
	\centering
	\centerline{\includegraphics[width=1\linewidth]{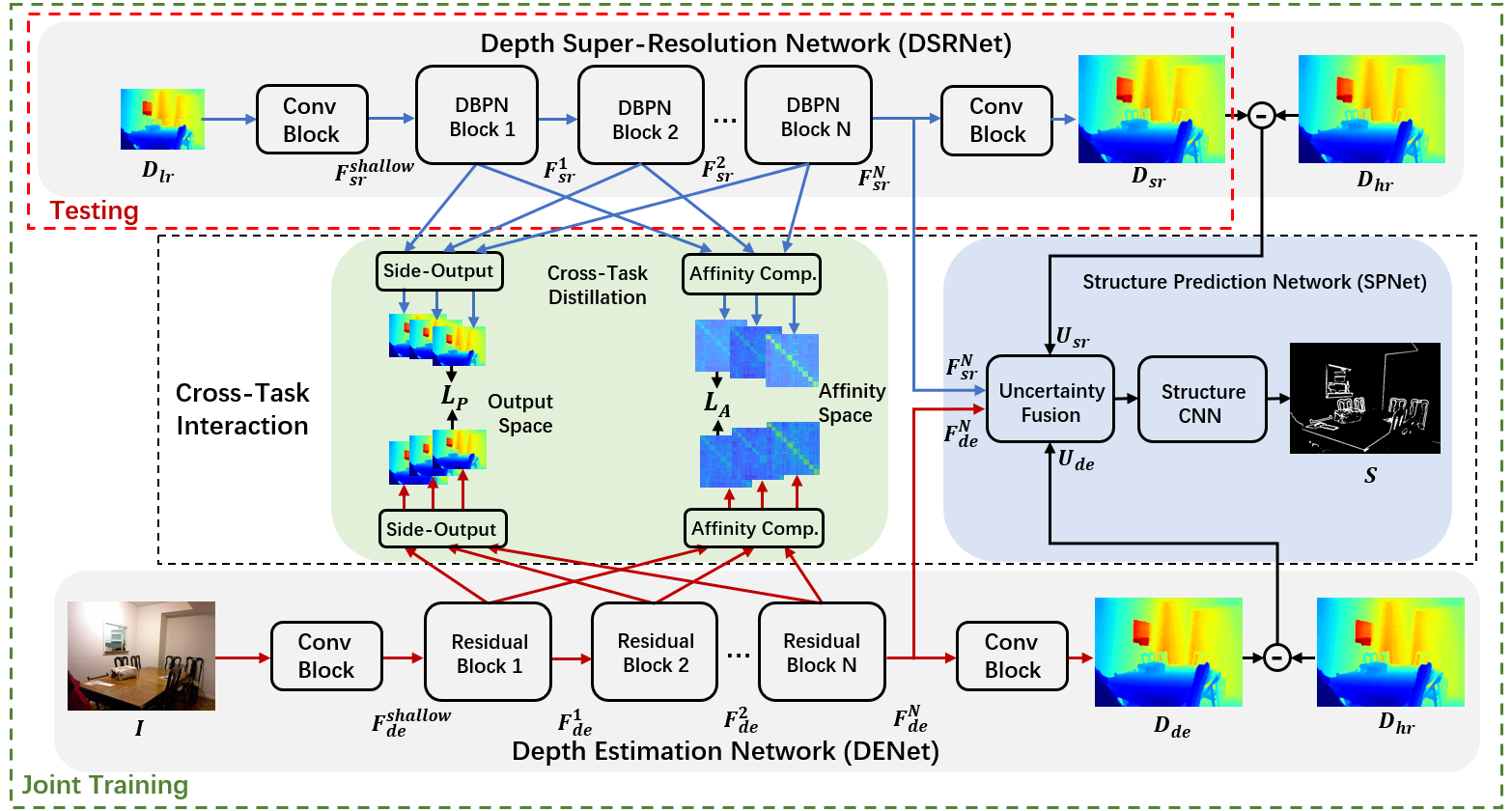}}
	\caption{Illustration of our proposed framework, which consists of DSRNet, DENet, and the middle cross-task interaction module. We supervise the outputs of DSRNet and DENet with the same groundtruth depth map $D_{hr}$.  In testing phase, DSRNet is the final choice to predict HR depth map from only LR depth map without the help of color image.}
	\label{fig:mainnet}
\vspace{-4mm}
\end{figure*}

Specifically, as illustrated in Figure \ref{fig:mainnet}, inspired by the success of multi-task learning \cite{DBLP:conf/cvpr/ZhangXHL18, DBLP:conf/iccv/KunduLR19, DBLP:journals/corr/abs-2008-07816}, we construct an auxiliary depth estimation (DE) task that takes RGB image as input to estimate a depth map. Upon this, we  propose a cross-task interaction module to realize bilateral knowledge transfer between DSR task and DE task. Different from the commonly used distillation techniques \cite{DBLP:conf/iclr/UrbanGKAWMPRC17, DBLP:journals/corr/HintonVD15, DBLP:conf/iccv/KunduLR19}, we first design a
cross-task distillation that encourages DSR network (DSRNet) and DE network (DENet) to learn from each other, i.e., the roles of teacher and student will dynamically switch between both tasks based on their current performances on depth recovery in the iterative collaborative training. A multi-space distillation scheme is introduced to distill knowledge from the perspective of output and affinity spaces, which can better describe the essential structural characteristics of depth map. Moreover, to address the problem of RGB-D structure inconsistency, we construct a structure prediction (SP) task that provides extra structure regularization to help both DSRNet and DENet learn more informative structure representations for depth recovery. We come up with an uncertainty-induced attention fusion module to provide a reasonable input for the SP network (SPNet), in which the uncertainty maps acquired from both DSRNet and DENet are used to re-weight their features for strengthening effective structural knowledge. Extensive experiments demonstrate that our single DSR method even outperforms the color-guided DSR methods on benchmark datasets in terms of both accuracy and runtime. The main contributions  are summarized as follows,

$\bullet$ So far as we know, our proposed paradigm of DSR is the first work  that learns with multiple modalities as inputs at training stage, but tests on only single LR depth modality.

$\bullet$ A cross-task distillation scheme is proposed to encourage DSRNet and DENet to learn from each other  in a collaborative training mode.

$\bullet$ A structure prediction network is advanced to provide structure regularization for helping DSRNet resolve the problem of structural inconsistency.


\section{Related Work}

\textbf{Depth Super-Resolution.}
Compared to single DSR methods \cite{ferstlimage, DBLP:conf/eccv/RieglerRB16, DBLP:journals/tmm/XieFYS15, DBLP:journals/tcsv/SongDQ19}, color-guided DSR methods \cite{DBLP:conf/eccv/HuiLT16, DBLP:journals/tip/GuoLGCFH19, WANG2020107274, DBLP:journals/tip/YeSWYXLL20} have been widely proposed to improve the quality of depth map by the guidance of color image. Li \emph{et al.} \cite{DBLP:conf/eccv/LiHA016} proposed a joint filtering approach that leverages color image as guidance to enhance the spatial resolution of depth map. Hui \emph{et al.} \cite{DBLP:conf/eccv/HuiLT16} employed a multi-scale fusion strategy that fuses the rich hierarchical color features at different levels to resolve ambiguity in DSR. Wen \emph{et al.} \cite{DBLP:journals/tip/WenSLLF19} presented a data-driven filter to infer an initial HR depth map with the guidance of color image, then proposed a coarse-to-fine network to progressively recover the depth map. Guo \emph{et al.} \cite{DBLP:journals/tip/GuoLGCFH19} proposed a hierarchical feature driven method that constructs an input pyramid and a guidance pyramid for multi-level residual learning. 
Wang \emph{et al.} \cite{WANG2020107274} proposed to upsample the depth map with the help of edge map learned from the color image.

\textbf{Monocular Depth Estimation.}
Due to the strong ability of CNN in  feature extraction, many supervised monocular depth estimation methods \cite{DBLP:conf/cvpr/0002000S018, DBLP:conf/cvpr/GodardAB17, DBLP:conf/cvpr/WongS19, DBLP:conf/cvpr/KunduUPB18} continue to improve the performance of depth estimation.  Laina \emph{et al.} \cite{DBLP:conf/3dim/LainaRBTN16} proposed a fully convolutional architecture to model the mapping between color image and depth map. In \cite{DBLP:journals/cviu/LiYKY19}, a two-stream CNN is proposed to simultaneously predict depth and depth gradients  for accurate depth estimation. Wang \emph{et al.} \cite{DBLP:conf/cvpr/WangZWLL20} presented a depth estimation network via a semantic divide-and-conquer strategy, in which a scene is decomposed into semantic segments and then predicts  depth  for each segment. In contrast, unsupervised methods  \cite{DBLP:conf/eccv/GargKC016, DBLP:conf/cvpr/GodardAB17, DBLP:conf/cvpr/WongS19, DBLP:conf/cvpr/ZhanGWLA018} use video or stereo data during training without the need of groundtruth depth maps. Wong \emph{et al.} \cite{DBLP:conf/cvpr/WongS19} learned a robust representation with a two-branch decoder to estimate the depth map. 
Different from the above works, we introduce an auxiliary monocular depth estimation task for achieving the cross-model knowledge learning at training stage.

\textbf{Knowledge Distillation.}
Knowledge distillation \cite{DBLP:journals/corr/HintonVD15} is to transfer knowledge from high-capacity model to a compact model to improve the performance of the latter one.  It has been wildly applied to many applications, including action recognition \cite{DBLP:conf/eccv/GarciaMM18}, style transfer \cite{DBLP:conf/ijcai/LiWLH17}, depth estimation \cite{DBLP:conf/eccv/GuoLYRW18} and scene parsing \cite{DBLP:conf/cvpr/0002OWS18}. For example, the task of image classification takes  class probabilities from teacher network as soft targets to train the student network \cite{DBLP:conf/iclr/UrbanGKAWMPRC17, DBLP:journals/corr/HintonVD15} or transfers the knowledge through intermediate layers \cite{DBLP:conf/iclr/ZagoruykoK17, DBLP:journals/corr/RomeroBKCGB14}.
Recently, deep mutual learning \cite{DBLP:conf/cvpr/ZhangXHL18} proposed a two-way distillation which transfers knowledge  between the teacher and student and benefits to both networks. Kundu \emph{et al.} \cite{DBLP:conf/iccv/KunduLR19} tried to extend cross-model distillation to multiple spatially-structured prediction tasks by using two regularization strategies to minimize the domain discrepancy.  Yao \emph{et al.} \cite{DBLP:journals/corr/abs-2008-07816} presented a dense cross-layer mutual distillation mechanism to train the teacher and student collaboratively from scratch. Inspired by the above knowledge distillation techniques,
we propose to learn the scene structure guidance for the single DSR task via our designed cross-task distillation. Moreover, another difference to the previous works is that the role of teacher and student is dynamically changed.

\section{Method}
\subsection{Network Architecture}

Figure \ref{fig:mainnet} shows the overall architecture of our framework, which mainly consists of three components: depth super-resolution network (DSRNet), depth estimation network (DENet) and the middle cross-task interaction module. Given a collection of paired LR-HR depth maps $\{D_{lr}^{(k)}, D_{hr}^{(k)}\}_{k=1}^{M}$ with the corresponding HR color images $\{I^{(k)}\}_{k=1}^{M}$ as  training data,  where $M$ is the number of training data, our goal is to learn a model, i.e., DSRNet, that can predict the super-resolved depth maps $\{D_{sr}^{(k)}\}_{k=1}^{M}$ from their corresponding downsampled versions $\{D_{lr}^{(k)}\}_{k=1}^{M}$.

Specifically, the structure of DSRNet is designed based on a network unit from deep back-projection network (DBPN) \cite{DBLP:conf/cvpr/HarisSU18}, which can effectively improve the feature representations through iterative projecting HR representations to LR spatial domain and then mapping back into HR domain\footnote{We direct readers to refer to \cite{DBLP:conf/cvpr/HarisSU18} for more details about the design of DBPN block. }. The shallow features $F_{sr}^{shallow}$  are first extracted from $D_{lr}$ through a simple convolutional block (including three convolutional layers), and then sent into $N$ stacked DBPN blocks to obtain HR features $\{F_{sr}^n\}_{n=1}^{N}$. $D_{sr}$ is finally reconstructed from $F_{sr}^N$ through another convolutional block. DENet takes $I$ as input and estimates the depth map $D_{de}$. The architecture of DENet is similar to DSRNet, but replaces the DBPN blocks with deeper residual blocks ~\cite{DBLP:conf/cvpr/HeZRS16} to extract informative features $\{F_{de}^n\}_{n=1}^{N}$ from color image.

The cross-task interaction module acts as a bridge to connect DSRNet and DENet, and realizes bilateral knowledge transfer between them. It consists of two components, i.e., a cross-task distillation scheme and a structure prediction network (SPNet), where the former focuses on the interaction between multi-scale features extracted from both networks while the latter uses structure maps as supervision to further guide the learning of both networks.

Note that, at the training stage, DSRNet, DENet and the cross-task interaction module are jointly learned by using both color image and depth map as input. In testing phase, DSRNet is the final choice to predict HR depth map from only LR depth map  without the help of color image.

\subsection{Cross-Task Distillation}
\label{secdistll}
Knowledge distillation is generally viewed as a technique of transferring beneficial information from a top-performing model to the other naive one. Different from the commonly used distillation techniques, in which the teacher network is trained beforehand and fixed under the assumption that it always learns a
better representation than the student network, our goal is to train SRNet and DENet collaboratively and encourage them to benefit from each other.

Inspired by mutual learning methods \cite{DBLP:conf/cvpr/ZhangXHL18, DBLP:conf/iccv/KunduLR19}, we propose a cross-task distillation scheme, in which the roles of teacher and student will exchange between both tasks based on their
current performances on depth recovery  in the iterative collaborative training. Specially, at the current round of training, we need to determine the teacher in advance according to their performance at the previous round. We compute the average pixel error between each recovered depth map and its groundtruth for both networks:
\begin{equation}\label{esr}
e_{dsr} = \frac{1}{HW}\sum_h^H\sum_w^W|D_{sr}(h,w) -D_{hr}(h,w)|,
\end{equation}
\begin{equation}\label{ede}
e_{de} = \frac{1}{HW}\sum_h^H\sum_w^W|D_{de}(h,w) -D_{hr}(h,w)|,
\end{equation}
where $\{H,W\}$ are the size of the output depth map. If $e_{dsr}$ is smaller than $e_{de}$, DSRNet has a relatively better performance, and becomes the dominant one to guide the learning of DENet, and vice versa.

Next, in order to distill more meaningful knowledge that can accurately describe the essential structural characteristics of depth map, we introduce a multi-space distillation scheme to condense the knowledge from the perspectives of output and affinity spaces, as shown in Figure \ref{fig:mainnet}.

\noindent \textbf{Output Space Distillation.} To ensure the transfer of local information from pixel-wise depth values in a depth map, we apply the side-output layer (containing two successive convolutions) on $\{F_{sr}^n, F_{de}^n\}_{n=1}^{N}$ from both DSRNet and DENet to generate the corresponding multi-scale depth outputs $\{D_{sr}^n, D_{de}^n\}_{n=1}^{N}$ respectively. Thus, the distillation loss of output space is designed to indirectly align the features between DSRNet and DENet:
\begin{equation}\label{d1}
\mathcal{L}_{O} = \frac{1}{N} \sum_{i=1}^N||D_{sr}^i - D_{de}^i||_1,
\end{equation}
\noindent \textbf{Affinity Space Distillation.} Color image and its associated depth map are different representations of the same scene and have strong structural similarity. Pixels with similar appearances in a color image have more chances of belonging to the same object, and should have close depth values. Inspired by \cite{DBLP:conf/cvpr/0004GGH18,2018Monocular,2020DPNet} that consider the nonlocal correlations to strengthen correlated features between pixels and benefit the depth map recovery, we also transfer non-local structure knowledge on affinity space, which is implemented by computing pair-wise similarities between pixels.

Assuming the dimension of feature $F$ is $w \times h \times c$, the reshape function $\mathbb{R}$ recasts $F$ as $\mathbb{R}(F)$ with the dimension of $wh \times c$. The affinity matrix $A$ is defined as:
\begin{equation}\label{d2}
A(F) = \sigma(\mathbb{R}(F)\otimes \mathbb{R}^{T}(F)),
\end{equation}
where $\sigma(\cdot)$ is the softmax operation, $\otimes$ is the matrix multiplication and $T$ is the transpose operator. The distillation loss of affinity space is defined as the following,
\begin{equation}\label{d3}
\mathcal{L}_{A} = \frac{1}{N} \sum_{i=1}^N||A(F_{sr}^i) - A(F_{de}^i)||_1,
\end{equation}

The final distillation loss $\mathcal{L}_{distill}$ is expressed as:
\begin{equation}\label{d4}
\mathcal{L}_{distill} = \mathcal{L}_{O} + \gamma \mathcal{L}_{A}.
\end{equation}
where $\gamma$ is an adjustment parameter. Note that, $\mathcal{L}_{distill}$ should be imposed on the training of the student, but not the teacher, which are determined by the errors comparison between  $e_{dsr}$ and $e_{de}$.

\subsection{Structure Prediction}

\begin{figure}[!t]
	\centering
	\centerline{\includegraphics[width=1\linewidth]{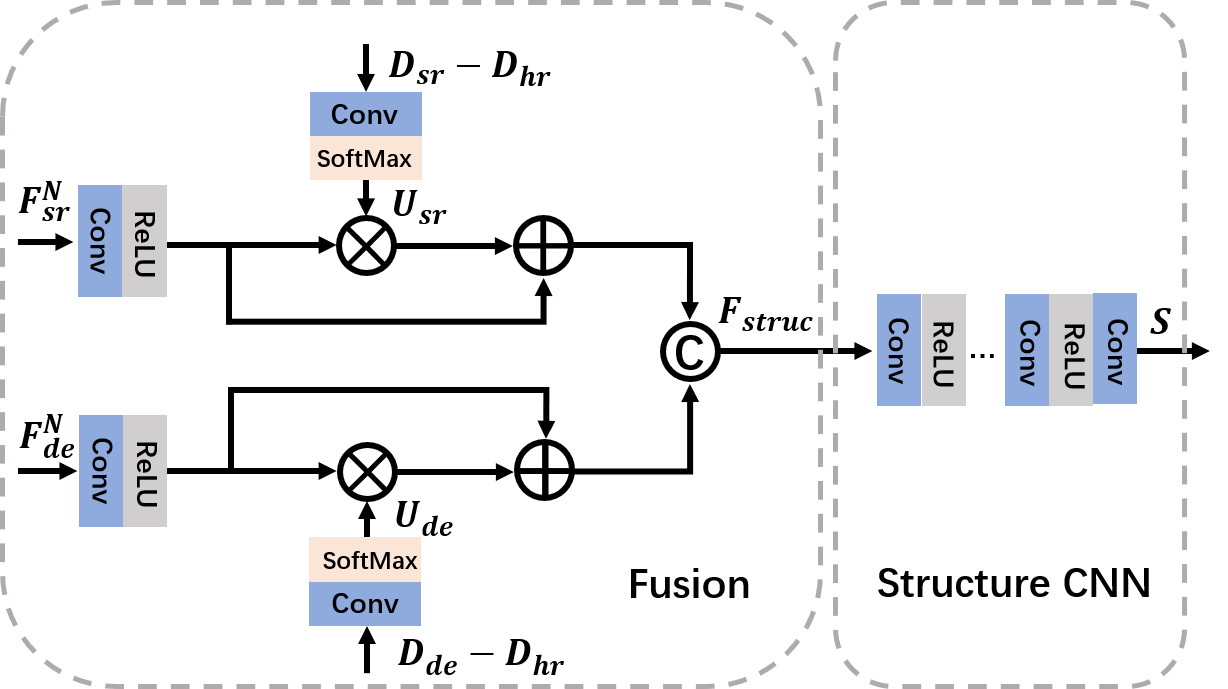}}
	\caption{The proposed SPNet. We fuse $F_{sr}^N$ and $F_{de}^N$ by our proposed uncertainty-induced attention fusion module.}
	\label{fig:grad}
\vspace{-4mm}
\end{figure}
The goal of SPNet is to predict a structure map $S$ from the last feature maps $F_{sr}^N$ and $F_{de}^N$ generated by DSRNet and DENet respectively. Through the supervision with the groundtruth structure map $S_{gt}$\footnote{$S_{gt}$ is obtained by computing the difference between adjacent pixels (gradients) in $D_{HR}$. Following~\cite{DBLP:conf/cvpr/MaRCCL020}, we calculate it by a convolution layer with a fixed kernel to extract the high-frequency parts from $D_{HR}$.}, SPNet can provide extra structure regularization to help both DSRNet and DENet learn more informative structure representations to alleviate the problem of RGB-D structure inconsistency. As shown in Figure \ref{fig:grad}, SPNet consists of a fusion module and a structure CNN, where the latter is a lightweight network with five stacked  `Conv+ReLU' layers and a last 'Conv' layer.

Usually, the erroneous recovery of DSR and DE tasks occurs in the regions around depth boundaries and fine structures in a depth map, which are subject to higher recovery uncertainty. Therefore, instead of simply concatenating $F_{sr}^N$ and $F_{de}^N$ and sending into the structure CNN, we design an uncertainty-induced attention fusion module to strengthen these informative structure features by attending the recovery uncertainty  to the feature map. Thus, we first compute the uncertainty maps $U_{sr}$ and $U_{de}$ of both networks by activating the recovery errors:
\begin{equation}\label{un2}
U_{sr} = \sigma(Conv_{1\times1}(D_{sr} - D_{hr}))
\end{equation}
\begin{equation}\label{un1}
U_{de} = \sigma(Conv_{1\times1}(D_{de} - D_{hr}))
\end{equation}
where $Conv_{1\times1}$ is the $1\times1$ convolution to adjust the channels. Then,
we use the uncertainty maps to re-weight $F_{sr}^N$, $F_{de}^N$ and fuse them through an attention module:
\begin{equation}
\label{grad1}
F_{struc} = [ F_{sr}^N * (1+U_{sr}),  F_{de}^N * (1+U_{de})]
\end{equation}
where $F_{struc}$ is the fused features for structure prediction. [$\cdot$] denotes concatenation operation and $*$ is element-wise multiplication. Note that, through back-propagating the network gradients from SPNet in the backward information flow, the parameters of DSRNet and DENet can be updated.

\begin{algorithm}[!t]
	\caption{Training Details}
	\label{algorithm11}
	\begin{algorithmic}[1]
		
		\Require Training data {$D_{lr}$, $D_{hr}$, $I$,  $S_{gt}$} 
        \State ---------------- Step 1 ----------------
        \State Randomly initialize DSRNet and DENet
        \For {i = 1; i $\le$ 100}
        \State Train DSRNet and DENet with $\mathcal{L}_{DSR}$ and $\mathcal{L}_{DE}$, respectively
		\EndFor
        \State ---------------- Step 2 ----------------
        \State Randomly initialize SPNet
		\For {i = 101; i $\le$ max epoch}
        \State Compute the average error value $e_{dsr}$ and $e_{de}$ according to Eq.(\ref{esr}) and Eq.(\ref{ede})
        \If {$e_{dsr} \le e_{de}$}
        \State Fix DSRNet and update DENet with
		\State $\mathcal{L} = \mathcal{L}_{DE} + \rho_1\mathcal{L}_{struc} + \rho_2\mathcal{L}_{distill}$
        \Else
        \State Fix  DENet and update DSRNet with
        \State $\mathcal{L} = \mathcal{L}_{DSR} + \rho_1\mathcal{L}_{struc} + \rho_2\mathcal{L}_{distill}$
        \EndIf
		\EndFor
        \Ensure $D_{sr}$
			
	\end{algorithmic}
\end{algorithm}

\subsection{Training Algorithm}
The training process of our framework can be divided into two steps, as presented in Algorithm \ref{algorithm11}. First, we separately train DSRNet and DENet with the groundtruth $D_{hr}$. The losses are defined as follows:
\begin{equation}\label{loss1}
\mathcal{L}_{DSR} = ||D_{sr} - D_{hr}||_1,
\end{equation}
\begin{equation}\label{loss2}
\mathcal{L}_{DE} = \lambda\frac{1-SSIM(D_{de}, D_{hr})}{2} + (1-\lambda)||D_{de}-D_{hr}||_1,
\end{equation}
where $\mathcal{L}_{DSR}$ is a common pixel-wise L1 loss for the task of DSR. Following \cite{DBLP:conf/cvpr/GodardAB17},  $\mathcal{L}_{DE}$ is set as a combination of the reconstruction loss (L1 loss) and structural similarity (SSIM). $\lambda$ is an adjustment parameter.

Then, we introduce the cross-task distillation between both networks with the loss $\mathcal{L}_{distill}$ in Eq.~(\ref{d4}).
At the same time, we randomly initialize SPNet, and train it together with DSRNet and DENet. The loss for SPNet is defined as:
\begin{equation}\label{loss3}
\mathcal{L}_{struc} = ||\mathbb{G}(F_{struc}) - S_{gt}||_1,
\end{equation}
where $\mathbb{G}(\cdot)$ denotes SPNet, $F_{struc}$ is the fused structure feature in Eq.(\ref{grad1}) and $S_{gt}$ is the ground truth of the structure.  If DSRNet is chosen as the student, the parameters of DENet are fixed at the current epoch, and DSRNet is updated  with the following loss:
\begin{equation}\label{loss4}
\mathcal{L} = \mathcal{L}_{DSR} + \rho_1\mathcal{L}_{struc} + \rho_2\mathcal{L}_{distill},
\end{equation}
where $\rho_1, \rho_2$ are the trade-off parameters. Otherwise, DSRNet is fixed, and DENet is updated with the loss $\mathcal{L}$ by replacing $\mathcal{L}_{DSR}$ with $\mathcal{L}_{DE}$.

\begin{table*}[!t]
	\begin{center}
		\caption{Quantitative DSR results (in MAD) on Middlebury 2005 dataset. `DSRNet w/o CT' and `DSRNet' denote the results of the proposed method without and with cross-task interaction scheme, respectively.}
		 \fontsize{8.0}{13}\selectfont
		\label{table:compar}
		\scalebox{0.70}{
			\begin{tabular}{c|cccc|cccc|cccc|cccc|cccc|cccc}				
				\toprule
				\multirow{2}[3]{*}{} & \multicolumn{4}{c}{\textit{Art}} & \multicolumn{4}{c}{\textit{Books}} & \multicolumn{4}{c}{\textit{Dolls}} & \multicolumn{4}{c}{\textit{Laundry}} & \multicolumn{4}{c}{\textit{Moebius}} & \multicolumn{4}{c}{\textit{Reindeer}}\\			\cmidrule{2-25}&$\times$2&$\times$4&$\times$8&$\times$16&$\times$2&$\times$4&$\times$8&$\times$16&$\times$2&$\times$4&$\times$8&$\times$16&$\times$2&$\times$4&$\times$8&$\times$16&$\times$2&$\times$4&$\times$8&$\times$16 &$\times$2&$\times$4&$\times$8&$\times$16 \\\hline
				Bicubic & 0.48  & 0.97  & 1.85  & 3.59  & 0.13  & 0.29  & 0.59  & 1.15  & 0.20   & 0.36  & 0.66  & 1.18 & 0.28  & 0.54  & 1.04  & 1.95  & 0.13  & 0.30   & 0.59  & 1.13  & 0.30   & 0.55  & 0.99  & 1.88\\
				DJF \cite{DBLP:conf/eccv/LiHA016}& 0.12  & 0.40   & 1.07  & 2.78  & 0.05  & 0.16  & 0.45  & 1.00     & \textbf{0.06}  & 0.20   & 0.49  & 0.99  & 0.07  & 0.28  & 0.71  & 1.67  & 0.06  & 0.18  & 0.46  & 1.02  & 0.07  & 0.23  & 0.60   & 1.36\\
				
				MSG \cite{DBLP:conf/eccv/HuiLT16}& \quad-  & 0.46  & 0.76  & 1.53  & \quad-  & 0.15  & 0.41  & 0.76  & \quad-  & 0.25  & 0.51  & 0.87 &  \quad-     & 0.30   & 0.46  & 1.12  & \quad-  & 0.21  & 0.43  & 0.76  & \quad-  & 0.31  & 0.52  & 0.99\\
				DGDIE \cite{DBLP:conf/cvpr/GuZGCCZ17}& 0.20   & 0.48  & 1.20   & 2.44  & 0.14  & 0.30   & 0.58  & 1.02  & 0.16  & 0.34  & 0.63  & 0.93 & 0.15  & 0.35  & 0.86  & 1.56  & 0.14  & 0.28  & 0.58  & 0.98  & 0.16  & 0.35  & 0.73  & 1.29\\
				DEIN \cite{DBLP:conf/icassp/YeDL18}&0.23   &0.40   &0.64   &\textbf{1.34}  &0.12   &0.22   &0.37   &0.78   &0.12   &0.22   &0.38   &0.73  &0.13   &0.23   &0.36   &0.81   &0.11   &0.20   &0.35  &0.73   &0.15   &0.26   &0.40  & 0.80 \\				
                CCFN \cite{DBLP:journals/tip/WenSLLF19}& \quad-  & 0.43  & 0.72  & 1.50   & \quad-  & 0.17  & 0.36  & 0.69  & \quad-  & 0.25  & 0.46  & 0.75 &   \quad-    & 0.24  & 0.41  & \textbf{0.71}  & \quad-  & 0.23  & 0.39  & 0.73  & \quad-  & 0.29  & 0.46  & 0.95\\
                GSRPT \cite{DBLP:conf/iccv/LutioDWS19}&0.22 &0.48 &0.74 &1.48&0.11 &0.21  & 0.38  &0.76&0.13 &0.28  & 0.48 & 0.79&0.12 &0.33  & 0.56  &1.24&0.12 & 0.24  &0.49  &0.80
                &0.14 & 0.31  &0.61  &1.07 \\
               DSRN \cite{WANG2020107274}&0.12   &0.25   &0.61   &1.80  &\textbf{0.04}   &\textbf{0.11}  &0.28   &0.69   &\textbf{0.06}   &\textbf{0.14}  &\textbf{0.33}   &0.73  &\textbf{0.06}   &\textbf{0.15}   &0.43   &1.24  &\textbf{0.05}   &\textbf{0.13}  &0.29  &\textbf{0.67}   &\textbf{0.07}   &\textbf{0.15}   &\textbf{0.35}  & 0.92 \\ \hline
               DSRNet w/o CT &0.16 &0.31 & 0.59&1.55 & 0.10& 0.15&  0.31 &0.73 &0.12 & 0.21& 0.39 &0.69 &0.12 & 0.21& 0.40  &0.82 & 0.11& 0.16& 0.32 &0.74 & 0.13&  0.22& 0.38 &0.87\\
				DSRNet & \textbf{0.11} & \textbf{0.25} & \textbf{0.53}& 1.44 & 0.05 & \textbf{0.11} & \textbf{0.26} & \textbf{0.67} &  0.07 & 0.16 & 0.36 & \textbf{0.65} & \textbf{0.06} & 0.16 & \textbf{0.36}  & 0.76 & 0.07 & \textbf{0.13} & \textbf{0.27} & 0.69 & 0.08 & 0.17 & \textbf{0.35} & \textbf{0.77} \\
				\bottomrule				
		   \end{tabular}}
	\end{center}
\vspace{-4mm}
\end{table*}

\begin{figure*}[ht]
	\centering
	\centerline{\includegraphics[width=1\linewidth]{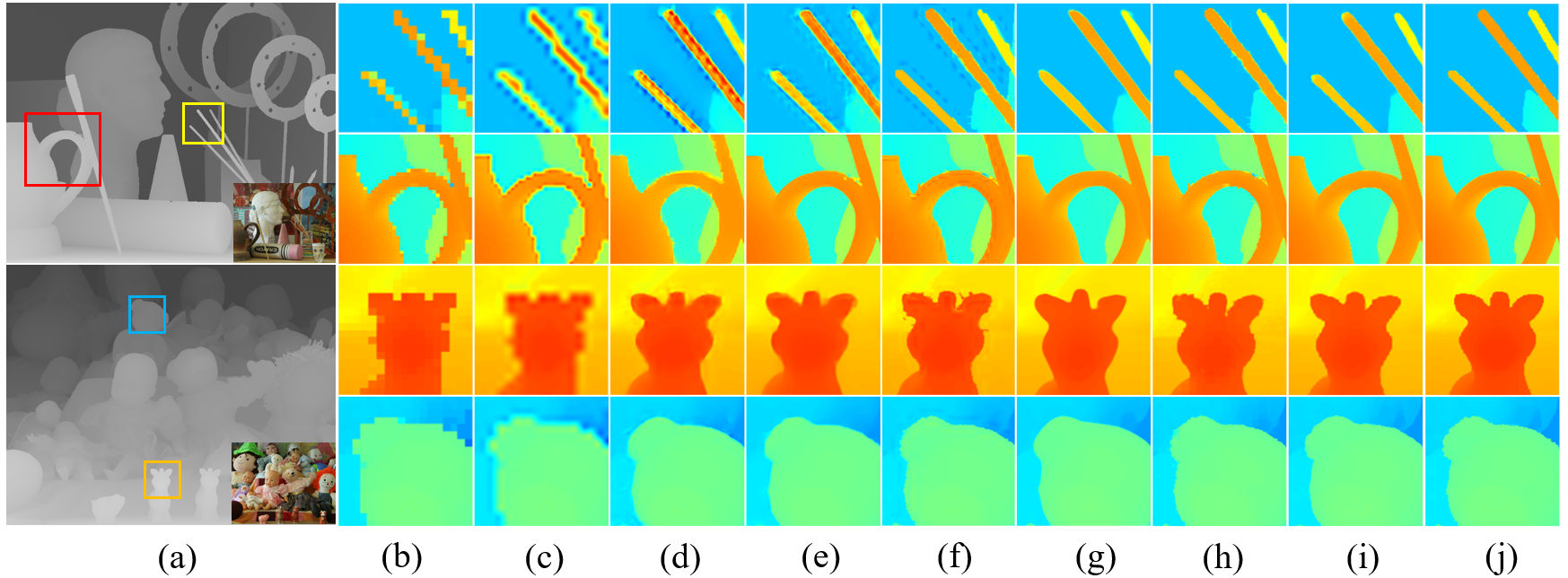}}
	\caption{Visual comparison of $\times$8 DSR results on \textit{Art} and \textit{Dolls} in Middlebury: (a) Groundtruth depth maps,  (b) LR patches, results from (c) Bicubic, (d) DJF \cite{DBLP:conf/eccv/LiHA016}, (e) DGDIE \cite{DBLP:conf/cvpr/GuZGCCZ17}, (f) DEIN \cite{DBLP:conf/icassp/YeDL18}, (g) GSRPT \cite{DBLP:conf/iccv/LutioDWS19}, (h) DSRN \cite{WANG2020107274}, (i) Ours, and (j) Groundtruth. }
	\label{fig:compar}
\end{figure*}

\section{Experiments}
\vspace{-8pt}
We conduct experiments on four datasets, i.e., Middlebury \cite{DBLP:conf/cvpr/HirschmullerS07}, MPI Sintel \cite{Butler:ECCV:2012}, NYUv2 \cite{DBLP:conf/eccv/SilbermanHKF12} and ToFMark \cite{ferstlimage}.
First, following MSG \cite{DBLP:conf/eccv/HuiLT16}, we use 34 RGB-D images from Middlebury dataset \cite{DBLP:conf/cvpr/HirschmullerS07} and 58 RGB-D images from MPI Sintel dataset \cite{Butler:ECCV:2012} as training data. To evaluate our performance, we test on Middlebury 2005 (6 standard RGB-D images: \emph{Art, Books, Moebius, Dolls, Laundry and Reindeer}). We also evaluate the generalization performance on ToFMark dataset (3 real-world depth maps captured by Time of Flight (ToF) sensor). Another training and testing dataset is NYU v2 dataset captured by Kinect sensor. Following the widely used data splitting manner, we use the first 1000 RGB-D images for training and the rest 449 RGB-D images for evaluation. For both experiment settings, we randomly extract 10000+ patches of fixed size of 256$\times$256 from HR depth maps and downsample HR depth maps by bicubic interpolation to get LR inputs. We augment the training data by $180^{\circ}$ rotation. We choose Mean Absolute Difference (MAD) and Root Mean Square Error (RMSE) as the evaluation metrics. Lower MAD and RMSE values indicate higher recovery quality.

During training, we set the number of DBPN and Residual blocks as $N=5$. We set the trade-off parameters as $\gamma = 0.5$, $\lambda = 0.2$, $\rho_1 = 0.1$ and $\rho_2 = 0.1$. For optimization, we use the Adam optimization algorithm with momentum $=0.9$, $\beta_1 = 0.9$, $\beta_2 = 0.99$ and $\epsilon = 10^{-8}$ to train our models. The initial learning rate is set to 1e-3 and decreased by multipling by 0.1 for every 50 epochs. We implemented our method using PyTorch with one RTX 2080Ti GPU.

\begin{figure*}[!t]
	\centering
	\centerline{\includegraphics[width=0.97\linewidth]{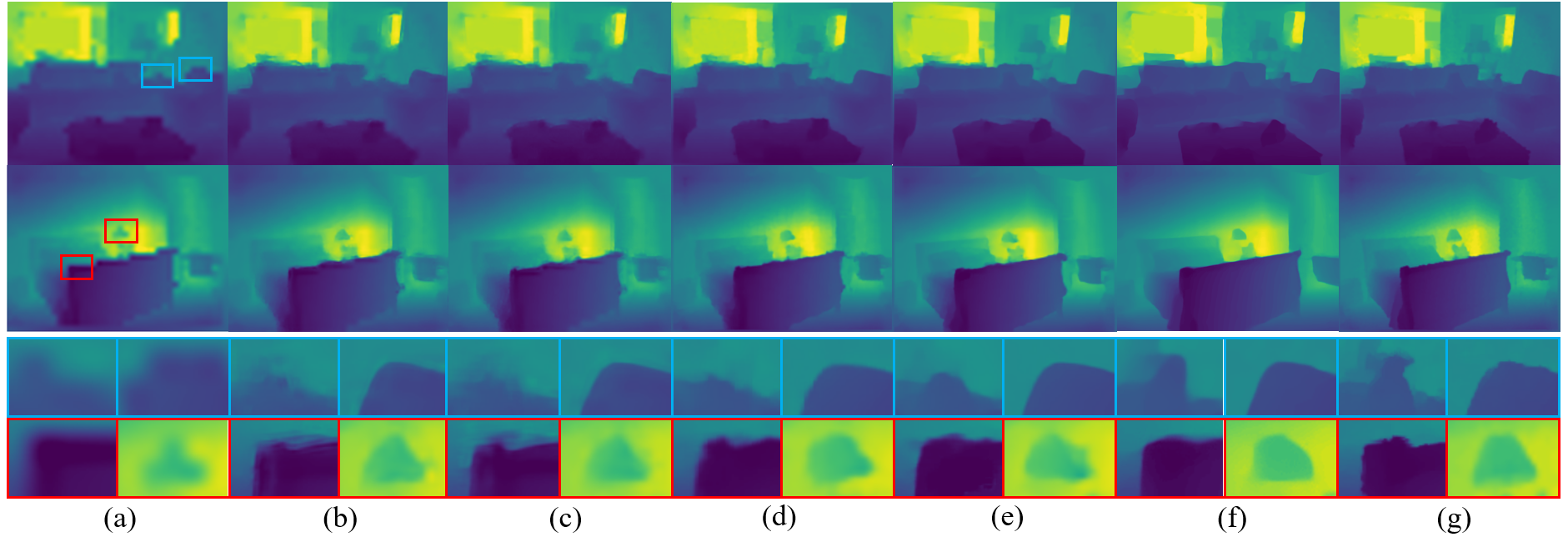}}
	\caption{Visual comparison of $\times$16 DSR results on NYU v2 dataset. (a) LR depth maps; Results from (b) DJF \cite{DBLP:conf/eccv/LiHA016}, (c) DJFR \cite{DBLP:journals/pami/LiHAY19}, (d) P2P \cite{DBLP:conf/cvpr/IsolaZZE17}, (e) GbFT \cite{DBLP:conf/iccv/AlbaharH19}, (f) Ours, and (g) Groundtruth.}
	\label{fig:nyu}
\vspace{-4mm}
\end{figure*}

\begin{figure}[!t]
	\centering
	\centerline{\includegraphics[width=1\linewidth]{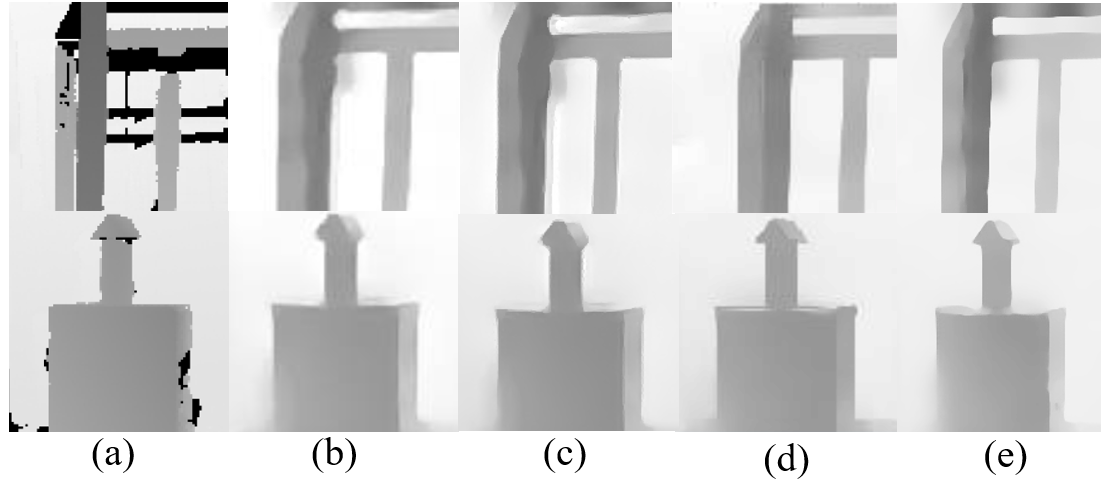}}
	\caption{Visualization on \emph{Shark} in ToFMark dataset. (a) Groundtruth patches, Results from (b) DGDIE\cite{DBLP:conf/cvpr/GuZGCCZ17}, (c)GSRPT\cite{DBLP:conf/iccv/LutioDWS19}, (d) DSRN\cite{WANG2020107274}, and (e) Ours.}
	\label{fig:tof}
\vspace{-2mm}
\end{figure}

\subsection{Comparison with State-of-The-Art}
\textbf{Middlebury Dataset.} We compare our method with state-of-the-art (SOTA) DSR methods on Middlebury  under four different up-scaling factors ($\times 2$, $\times 4$, $\times 8$ and $\times 16$).
All the compared methods and ours are deep learning based methods, which are trained and tested under the same conditions for fair comparison.
Table \ref{table:compar} summarizes the quantitative results on the Middlebury dataset.
`DSRNet w/o CT' and `DSRNet' denote the results of the proposed method without and with cross-task interaction scheme, respectively.
Benefiting from the backbone of DBPN network, the performance of `DSRNet w/o CT' goes beyond most of the previous methods, but slightly inferior to the recent SOTA methods, e.g., DSRN. When training together with DENet assisted by cross-task interaction, the performance of our DSRNet steps further compared to `DSRNet w/o CT', and is comparable to DSRN, even better than it on at least half of the cases. Note that, different from these color-guided methods, we stand on single DSR without the help of color image in testing phase, but achieves satisfactory results on both accuracy and runtime (evaluated later in this section).
The $\times$8 DSR results are visually shown in Figure \ref{fig:compar}. For the structural details, e.g., the stick and the teapot handle in \emph{Art}, and the toy's contour in \emph{Dolls}, we clearly recover these regions without introducing the texture copying artifacts thanks to our structure prediction network (SPNet).

\textbf{NYU Dataset.} We further evaluate our method on NYUv2 dataset compared with other SOTA methods in Table \ref{table:nyu}. Our method performs the best  on all the  cases with different up-scaling factors.
Figure \ref{fig:nyu} presents visual comparisons under the $\times$16 DSR case. Observing from the recovered depth maps and enlarged patches, our method achieves the  clearest results and preserves the sharpest and correct object contours,
which  demonstrates our effectiveness to learn the information of scene structure guidance from color image.

\textbf{ToFMark Dataset.} We evaluate the generalization ability of our method on ToFMark dataset. Following DGDIE \cite{DBLP:conf/cvpr/GuZGCCZ17}, we fill the missing points in depth maps and downsample them by $\times$2 downsampling factors, then send them to the $\times$2 model trained on Middlebury dataset to acquire the generalization results. As shown in Table \ref{table:real}, our method obtains the best objective generalization results for all three test examples. Figure \ref{fig:tof} shows the visual comparison on \emph{Shark}. Our method has higher visual quality and less blur than others, especially at the boundary regions.

\begin{figure*}[t]
	\centering
	\centerline{\includegraphics[width=1\linewidth]{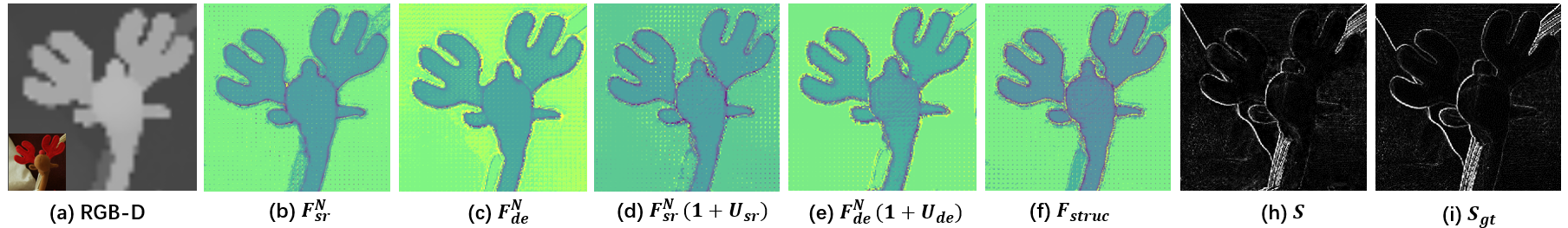}}
	\caption{Visual analysis of the fusion process. (a) RGB-D pairs, Output features from (b) DSRNet and (c) DENet, Re-weighted features  (d) and (e) corresponding to (b) and (c), respectively, (f) Fused features, (h) Predicted structure map; (i) Groundtruth structure map.}
	\label{fig:ab_f}
\vspace{-4mm}
\end{figure*}

\begin{figure}[!t]
	\centering
    \centerline{\includegraphics[width=1\linewidth]{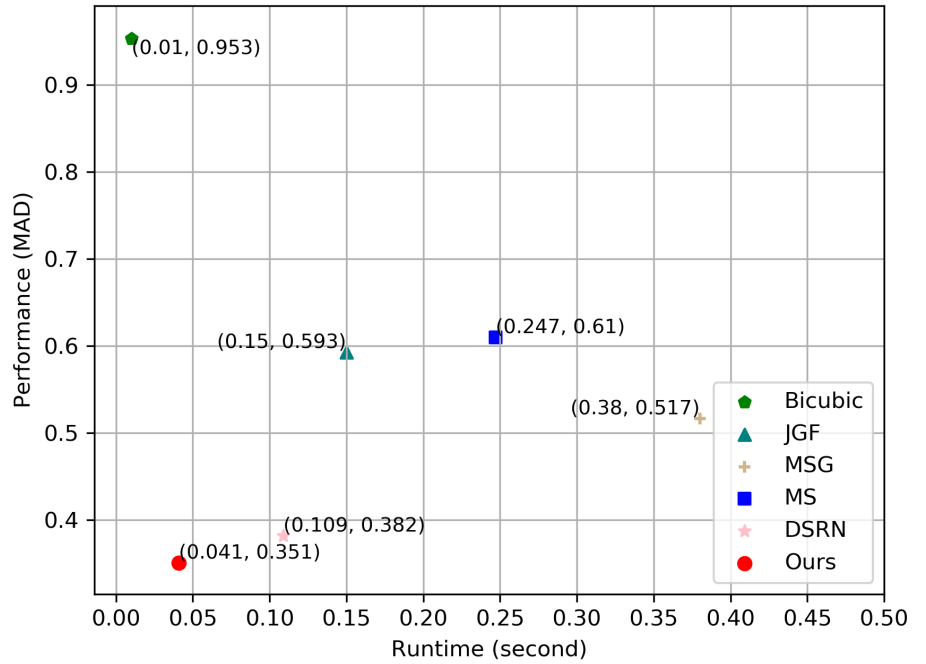}}
	\caption{Runtime and performance analysis.}
	\label{fig:run_times}
\vspace{-2mm}
\end{figure}

\textbf{Runtime.} In Figure \ref{fig:run_times}, we summarize the overall performance by the tradeoff between accuracy and runtime.
We measure the  runtime for $\times$8 DSR to their full resolution (about 1080$\times$1320) on Middlebury dataset. The color-guided methods i.e., JGF \cite{DBLP:conf/cvpr/0001TT13}  MSG \cite{DBLP:conf/eccv/HuiLT16}, and DSRN \cite{WANG2020107274}, run slower than ours due to the usage of HR color images in testing phase. MS is the single DSR version of MSG, but still inferior to ours for both accuracy and speed. Owing to the cross-task interaction, we achieve satisfactory recovered results with minimum inference time.

\begin{table}[!t]
	\begin{center}
		\caption{Quantitative DSR results (in RMSE) on NYUv2 dataset.}
		 \fontsize{8.0}{13}\selectfont
		\label{table:nyu}
		\scalebox{0.65}{
			\begin{tabular}{c|cccccccc}			
				\toprule
				\multirow{1}[1]{*}{}   & \multicolumn{1}{c}{DJF \cite{DBLP:conf/eccv/LiHA016}}&
				\multicolumn{1}{c}{DJFR \cite{DBLP:journals/pami/LiHAY19}}
				&\multicolumn{1}{c}{DGDIE \cite{DBLP:conf/cvpr/GuZGCCZ17}}  & \multicolumn{1}{c}{GbFT \cite{DBLP:conf/iccv/AlbaharH19}} &\multicolumn{1}{c}{P2P \cite{DBLP:conf/cvpr/IsolaZZE17}} &\multicolumn{1}{c}{PAC \cite{DBLP:conf/cvpr/SuJSGLK19}} &\multicolumn{1}{c}{DKN \cite{DBLP:journals/corr/abs-1903-11286}} & \multicolumn{1}{c}{Ours}\cr
				\hline
				$\times$4  & 3.54  & 3.38  &1.56&3.35&4.12&2.39&1.62&\textbf{1.49}\\
                $\times$8  & 6.20  & 5.86  &2.99&5.73&6.48&4.59&3.26&\textbf{2.73}  \\
                $\times$16 & 10.21  & 10.11 &5.24&9.01&10.17&8.09 &6.51&\textbf{5.11}\\
				\bottomrule				
		\end{tabular}}
	\end{center}
\vspace{-4mm}
\end{table}

\begin{table}[!t]
	\begin{center}
		\caption{Quantitative DSR results (in MAD) on ToFMark dataset.}
		 \fontsize{8.0}{13}\selectfont
		\label{table:real}
		\scalebox{0.79}{
			\begin{tabular}{c|cccccc}			
				\toprule
				\multirow{1}[1]{*}{} &\multicolumn{1}{c}{MSG \cite{DBLP:conf/eccv/HuiLT16}} &\multicolumn{1}{c}{DEIN\cite{DBLP:conf/icassp/YeDL18}}&\multicolumn{1}{c}{DGDIE \cite{DBLP:conf/cvpr/GuZGCCZ17}} &\multicolumn{1}{c}{GSRPT \cite{DBLP:conf/iccv/LutioDWS19}} & \multicolumn{1}{c}{DSRN \cite{WANG2020107274}}  & \multicolumn{1}{c}{Ours}\cr
				\hline
				\textit{Books}&12.26&12.78&12.31&13.21& 11.15&\textbf{11.03} \\
                \textit{Shark}&14.11&15.11&14.06&15.03& 13.26&\textbf{13.08} \\
                \textit{Devil}&12.45&14.25&9.66&12.27& 9.54&\textbf{9.33} \\
				\bottomrule				
		\end{tabular}}
	\end{center}
\vspace{-4mm}
\end{table}

\subsection{Ablation Study}
In this section, we further verify the key designs of our cross-task interaction, i.e., cross-task distillation and structure prediction task by ablation study.

\begin{table}[!t]

	\begin{center}
		\caption{The impact of each component in cross-task distillation.}
		 \fontsize{8.0}{13}\selectfont
		\label{table:ab2}
		\scalebox{0.85}{
			\begin{tabular}{cccccc|c|c}			
				\toprule
				\multirow{1}[1]{*}{}&DSRNet &DENet&$\mathcal{L}_{O}$&$\mathcal{L}_{A}$&SPNet& Middlebury & NYUv2\cr
				\hline
                &\checkmark&&&&&0.398&3.13\\
                &&\checkmark&&&&0.419&3.26\\ \hline
                &\checkmark&\checkmark&\checkmark&&&0.377&2.88\\
                &\checkmark&\checkmark& &\checkmark&&0.384&2.93\\
                &\checkmark&\checkmark&\checkmark&\checkmark&&0.372&2.85\\ \hline
                &\checkmark&\checkmark&\checkmark&\checkmark&\checkmark&\textbf{0.356}&\textbf{2.75}\\
				\bottomrule				
		\end{tabular}}
	\end{center}
\vspace{-4mm}
\end{table}

\begin{figure}[!t]
	\centering
	\centerline{\includegraphics[width=1\linewidth]{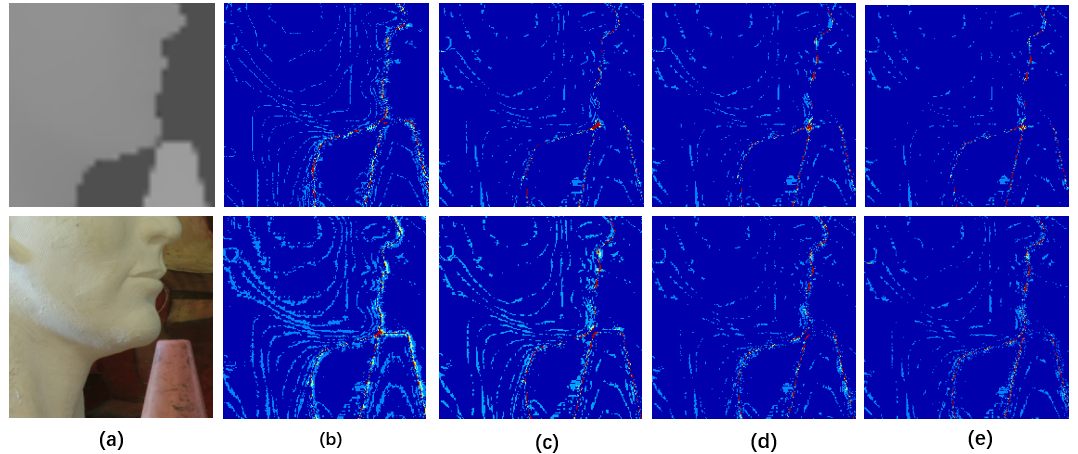}}
	\caption{Visualization of recovery errors variation in DSRNet at different distillation stage. (a) LR depth and HR RGB patches; (b) Error maps w/o cross-task distillation; Error maps with cross-task distillation after (c) 140$th$ and (d) 180$th$ training epoches; (e) Error maps with cross-task distillation and structure prediction.}
	\label{fig:ab_e}
\vspace{-6mm}
\end{figure}

We report the $\times$8 DSR results  on Middlebury and NYUv2 dataset under different experimental settings, as shown  in Table \ref{table:ab2}.
The $1^{st}$ and $2^{nd}$ rows are the objective results of initial DSRNet and DENet trained separately without cross-task interaction scheme. DENet performs relatively inferior to DSRNet, because estimating depth map from only color image is more difficult than from the degraded LR version. Next, $\mathcal{L}_{S}$ and $\mathcal{L}_{A}$ distill knowledge from  output space and affinity space respectively. When gradually integrating each loss into the baseline,  it brings a progressive performance improvement (6.5\% and 8.9\% increase on Middlebury and NYU respectively), which verifies the effectiveness of cross-task distillation strategy is remarkable. Besides, we also visualize the error maps between recovered depth map and groundtruth to validate the process of our iterative collaborative training in Figure \ref{fig:ab_e}. As the epoch goes, the recovery errors for both DSRNet and DENet are decreased at the boundary regions. Finally, after introducing the SPNet (the final case in Table \ref{table:ab2} ), the values of both error metrics  are further decreased (4.3\% error reduction against the previous case). Figure \ref{fig:ab_e}(e) also validates this from visual performance.

Besides, we conduct ablation study to evaluate the effectiveness of our uncertainty-induced attention fusion module as presented in Table \ref{table:ab_u}. Compared to simply concatenating features in channel dimension and sending into SPNet (SPNet w/o U), our proposed fusion module  can offer significant assistance for SPNet to predict the structure map.
We also visualize the feature maps in the fusion process in Figure \ref{fig:ab_f}. The object boundaries, e.g., the regions around the toy head and the strip behind, are obviously highlighted by the uncertainty map, which facilitate the final structure  perception of SPNet.

\begin{table}[!t]
	\begin{center}
		\caption{Ablation study of uncertainty-induced attention fusion module (abbreviated as U) on $\times$8 DSR cases. The numerical results represent the MADs computed between $S$ and $S_{gt}$. }
		 \fontsize{8.0}{13}\selectfont
		\label{table:ab_u}
		\scalebox{0.88}{
			\begin{tabular}{c|cccccc}			\toprule
            & \textit{Art} & \textit{Books} & \textit{Dolls} & \textit{Laundry} & \textit{Moebius} & \textit{Reindeer}\cr
            \hline

				SPNet w/o U &1.403 & 2.292 & 1.730 & 1.846 & 1.379& 1.565 \\
                SPNet w/ U &\textbf{1.386}&\textbf{2.259}&\textbf{1.715}&\textbf{1.821}& \textbf{1.349}&\textbf{1.552} \\
                \bottomrule				
		\end{tabular}}
	\end{center}
\vspace{-4mm}
\end{table}
\section{Conclusion}

For the first time, we explored to learn the cross-modal knowledge from both RGB and depth modalities at training stage, but test on only single depth modality. A cross-task interaction module is advanced to realize bilateral knowledge transfer between DSRNet and the auxiliary DENet in a well-designed collaborative training mode. Experiments show our method's superior performance for both accuracy and runtime. In the future work, we may extend our cross-task interaction  to more guided image restoration tasks.

{\small
\bibliographystyle{ieee_fullname}
\bibliography{ref}
}
\end{document}